\pdfoutput=1
\PassOptionsToPackage{dvipsnames}{xcolor}
\documentclass[11pt]{article}

\usepackage{acl}
\usepackage{amsthm}
\usepackage{nicefrac}
\usepackage{times}
\usepackage{latexsym}
\usepackage{graphicx} %
\usepackage{amsmath}
\usepackage{amssymb}
\usepackage[T1]{fontenc}
\usepackage{caption}
\usepackage{subcaption}
\usepackage[utf8]{inputenc}
\usepackage[dvipsnames]{xcolor}

\usepackage{microtype}
\usepackage{mathtools}
\usepackage{inconsolata}
\usepackage[capitalize,noabbrev]{cleveref}
\usepackage{pgfplots}
\usepackage{multirow}
\usepackage{tablefootnote}

\crefname{section}{Sec.}{Sec.}
\crefname{thm}{Thm.}{Theorem}
\crefname{appendix}{App.}{Appendices}
\crefname{algorithm}{Alg.}{Algorithms}
\crefname{equation}{Eq.}{Eqs.}
\crefname{figure}{Fig.}{Figs.}
\creflabelformat{equation}{#2\textup{#1}#3} %

\theoremstyle{plain}

\theoremstyle{definition}

\theoremstyle{remark}

\definecolor{tablecolor}{rgb}{0.8,0.8,0.8}

\newcommand{\vocab}{\mathcal{V}}

\newcommand{\vparam}{\vtheta}

\newcommand\cut[1]{}

\newcommand{\squishlist}{
   \begin{list}{$\bullet$}
    { \setlength{\itemsep}{0pt}      \setlength{\parsep}{3pt}
      \setlength{\topsep}{3pt}       \setlength{\partopsep}{0pt}
      \setlength{\leftmargin}{1.5em} \setlength{\labelwidth}{1em}
      \setlength{\labelsep}{0.5em} } }

\newcommand{\squishlisttwo}{
   \begin{list}{$\bullet$}
    { \setlength{\itemsep}{0pt}    \setlength{\parsep}{0pt}
      \setlength{\topsep}{0pt}     \setlength{\partopsep}{0pt}
      \setlength{\leftmargin}{2em} \setlength{\labelwidth}{1.5em}
      \setlength{\labelsep}{0.5em} } }

\newcommand{\squishend}{
    \end{list}  }

{}
{}
{}

\newcommand{\myvec}[1]{\mbox{$\mathbf{#1}$}}
\newcommand{\myvecsym}[1]{\mbox{$\boldsymbol{#1}$}}

\newcommand{\vtheta}{\mbox{$\myvecsym{\theta}$}}

\newcommand{\vy}{\mbox{$\myvec{y}$}}

\newcommand{\calD}{\mbox{${\cal D}$}}

\newcommand{\data}{\calD}

\title{Token Weighting for Long-Range Language Modeling}

\author{Falko Helm\hspace{10mm} Nico Daheim \hspace{10mm} Iryna Gurevych \\
  Ubiquitous Knowledge Processing Lab (UKP Lab) \\
Department of Computer Science and Hessian Center for AI (hessian.AI) \\
Technical University of Darmstadt \\
  \href{www.ukp.tu-darmstadt.de}{www.ukp.tu-darmstadt.de} \\
  }

\begin{document}
\maketitle

\begin{abstract}
Many applications of large language models (LLMs) require long-context understanding, but models continue to struggle with such tasks. 
We hypothesize that conventional next-token prediction training could contribute to this, because each token is assigned equal weight. 
Yet, intuitively, the amount of context needed to predict the next token accurately varies greatly across different data. 
To reflect this, we propose various novel token-weighting schemes that assign different weights to each training token in the loss, thereby generalizing existing works. 
For this, we categorize token-weighting methods using a two-step framework which compares the confidences of a long-context and short-context model to score tokens. 
We evaluate all methods on multiple long-context understanding tasks and show that non-uniform loss weights are helpful to improve the long-context abilities of LLMs.
Different short-context models can be used effectively for token scoring, including models that are much smaller than the long-context model that is trained.
All in all, this work contributes to a better understanding of the trade-offs long-context language modeling faces and provides guidelines for model steering via loss-weighting based on empirical evidence. 
The code can be found on \href{https://github.com/UKPLab/naacl2025-token-weighting}{Github}.

\end{abstract}

\definecolor{1g}{RGB}{255,255,255}
\definecolor{3g}{RGB}{213,255,213}
\definecolor{5g}{RGB}{171,255,171}
\definecolor{7g}{RGB}{129,255,129}
\definecolor{9g}{RGB}{89,255,89}
\definecolor{11g}{RGB}{47,255,47}
\definecolor{13g}{RGB}{5,255,5}

\definecolor{1r}{RGB}{255,255,255}
\definecolor{3r}{RGB}{255,213,213}
\definecolor{5r}{RGB}{255,171,171}
\definecolor{7r}{RGB}{255,129,129}
\definecolor{9r}{RGB}{255,89,89}
\definecolor{11r}{RGB}{255,47,47}
\definecolor{13r}{RGB}{255,5,5}

\begin{table*}[]
    \centering
    \resizebox{\linewidth}{!}{
    \begin{tabular}{l|ccccccccccccc}
    & \multicolumn{13}{c}{%
    {\huge "... inspired by the prints in one of Melquíades’ books ..."} } \\ \hline
    \begin{tabular}{c}Long-Context \\ Model Loss \end{tabular} &8.75	&0.20	&0.58	&7.82	&1.32	&4.54	&0.02	&3.76	&0.02	&<0.01	&0.01	&0.04	&0.26 \\ \hline 
    \begin{tabular}{c}Short-Context \\ Model Loss\end{tabular}& 8.49&0.21	&0.64	&8.82	&1.03	&4.40	&0.02	&4.84&0.63	&0.01	&0.00	&0.72	&0.46 \\ \hline
    \begin{tabular}{c}Absolute Loss \\ Difference\end{tabular}& 
    \colorbox{5g}{0.26}	&
    \colorbox{13r}{0.01}	&
    \colorbox{9r}{0.07}	&
    \colorbox{13g}{1.00}	&
    \colorbox{5g}{0.29}&	
    \colorbox{5r}{0.14}	&
    \colorbox{13r}{<0.01}	&
    \colorbox{13g}{1.08}&
    \colorbox{9g}{0.61}	&
    \colorbox{13r}{<0.01}	&
    \colorbox{13r}{0.01}	&
    \colorbox{9g}{0.68}	&
    \colorbox{3g}{0.20} \\ \hline
    \begin{tabular}{c}Loss \\ Weights \end{tabular}&  
    \begin{tabular}{c}{ \LARGE \colorbox{5g}{inspired}}\end{tabular} &
    \begin{tabular}{c}{\LARGE\colorbox{13r}{by}}\end{tabular} &
    \begin{tabular}{c}{\LARGE\colorbox{9r}{the}}\end{tabular} &
    \begin{tabular}{c}{\LARGE\colorbox{13g}{prints}}\end{tabular} &
    \begin{tabular}{c}{\LARGE\colorbox{5g}{in}}\end{tabular} &
    \begin{tabular}{c}{\LARGE\colorbox{5r}{one}}\end{tabular} &
    \begin{tabular}{c}{\LARGE\colorbox{13r}{of}}\end{tabular} &
    \begin{tabular}{c}{\LARGE\colorbox{13g}{Mel}}\end{tabular}&
    \begin{tabular}{c}{\LARGE\colorbox{9g}{qu}}\end{tabular}&
    \begin{tabular}{c}{\LARGE\colorbox{13r}{í}}\end{tabular}&
    \begin{tabular}{c}{\LARGE\colorbox{13r}{ades}}\end{tabular}&
    \begin{tabular}{c}{\LARGE\colorbox{9g}{’}}\end{tabular} &
    \begin{tabular}{c}{\LARGE\colorbox{3g}{books}}\end{tabular} 
    \end{tabular}
    }
    \caption{Schematic example of our token scoring method. Sequences are processed with either long and short context. \colorbox{13g}{Green}: weight bigger than one due to high absolute loss difference. \colorbox{13r}{Red}: weight smaller than one due to low absolute loss difference. The example sequence is taken from chapter 6 of "One Hundred Years of Solitude" \cite{marquez_one_2000}, where character \textit{Melquíades} reappears after having not been mentioned for 10k tokens before that passage. A model with 8k context sees this name for the first time and is uncertain. A 32k model can look back far enough to achieve a lower loss. Based on this, our method assigns a high weight to the "Mel" token, indicating a long-range dependency. Also, tokens which are learnable for the long-context model ("inspired") are upweighted, whereas trivial ("ades") and inherently hard ("one") tokens are downweighted.}
    \label{tab:Case-Study}
\end{table*}

\section{Introduction}
Many Natural Language Processing applications require models to reason about large amounts of contiguous texts, such as legal documents or textbooks in education applications~\citep{wang2024book2dial}.
To solve such tasks, recently various Large Language Models (LLMs) have been proposed that can process such large texts in one forward pass by allowing a large input context
 \cite{dubey_llama_2024}. 
However, while they make use of various approaches, such as data augmentation or different positional encodings, it is still unclear how well the resulting models are actually able to use their context \cite{liu2024lost} 
and what exactly contributes to making LLMs understand long contexts better. 

Broadly speaking, related works on training long-range LLMs tend to focus on two aspects, namely the training data and model architecture. 
Since long-range dependencies can be scarce, also because the number of possible strings explodes with sequence length and only a small share might be captured in common training data, model-based dataset filtering \cite{chen_long_2024} and data augmentation with synthetic long-range dependencies \cite{wu_long_2024} is often used to up-weigh long-range data. 
In terms of model architecture, there is a focus on computational efficiency, as the complexity of dot-product attention-based Transformers scales quadratically with the input length. 
For example, sub-quadratic attention mechanisms \cite[inter alia]{tay2022efficient} and Transformers with a memory mechanism, in which previous (compressed) context is stored, have been proposed \cite{wu2022memorizing, bulatov2022recurrent}. 
Finally, various positional encodings like RoPE \cite{su2024roformer} and ALiBi \cite{press_train_2022} are often used to facilitate long-context understanding. 
However, little attention has been paid to the link between model and data: the loss function and training criterion.

We hypothesize that the lack of long context capabilities could also come from a training criterion that weighs the contribution of each token equally. 
Intuitively, a weighting that emphasizes long-range dependencies (LRDs) by weighing data that exhibits such dependencies higher should be beneficial for model performance on these tasks. While there are existing approaches which use non-uniform loss weights~\citep{lin_rho-1_2024, ren_learning_2019}, they rather focus on efficiency and the specific choices that are made in them are not well understood.
In this work we generalize these methods in a comprehensive framework and thoroughly evaluate how token weightings should be chosen for better long-context performance in LLMs.

Our framework divides token weighting into two subsequent stages: scoring and postprocessing. 
Token scoring contrasts the confidences of a short- and a long-context model and sets them to dense or sparse weights. We compare these two types of postprocessing in detail. 
Furthermore, we compare using a frozen pretrained model for the short-context model (which can also be much smaller than the long-context model) against using the same long-context model with artificially shortened input context as a short-context model.
We pose the following two research questions: \begin{enumerate}
    \item What are the effects of sparse and dense weightings (i.e. of postprocessing)?
    \item What is the effect of the token scoring?
\end{enumerate}

Our experiments focus on extending the context of an existing language model~\citep{dubey_llama_2024, stallone_scaling_2024}.
In particular, we extend the context of Llama-3 8B and Phi-2 2.7B from 8k resp. 2k to 32k input tokens and compare different token weighting methods on RULER~\citep{hsieh_ruler_2024} and Longbench~\citep{bai_longbench_2023}, while contrasting how well they retain performance on short contexts as measured on MMLU~\citep{mmlu_hendrycks_21} and BBH \cite{suzgun_bbh}. 
Our results show that non-uniform token weights can improve the long-context performance of LLMs effectively and are better than using uniform weights.
This holds across various short-context models, namely, a frozen version of the long-context model, a weight-shared model with artificially shortened context, and a much smaller (8x) model, similar to weak-to-strong generalization~\citep{burns24weaktostrong}.
Still, improving long-context capabilities can form a trade-off with retaining original performance as shown in extensive ablations with different sparsity levels and interpolation strengths.

\section{Background}
\label{sec:background}
 In this section, we first introduce LLM training as next-token prediction with per-token weights (\cref{subsec: next-token prediction}).
 Then, we re-interpret it in terms of the training data, which allows token weighting to be understood as data curation (\cref{subsec:dataset curation}).
 Finally, we provide an overview of related works that use (non-uniform) token weights (\cref{subsec: loss weighting}).

\subsection{Next-Token Prediction}
\label{subsec: next-token prediction}
Current-day large language models (LLMs) are usually trained on large collections of text using a self-supervised next-token prediction objective \cite{radford2019language, touvron_llama_2023}.  
Corresponding to that, LLMs usually define a distribution over tokens in a sequence $\vy = (y_1, \ldots, y_N) \in \vocab^{N}$ with variable but bounded length $N$ that is constructed from a predetermined vocabulary $\vocab$, for example, of Byte-Pair-Encoding tokens~\citep{sennrich-etal-2016-neural}. 
Then, the probability of a sequence $\vy$ is determined autoregressively from a locally-normalized model as:
\begin{equation}
    p_{\text{\vparam}}(\vy) = \prod_{i=1}^N p_{\text{\vparam}}(y_i\mid \vy_{<i}).
\end{equation}
Here, $\vparam$ are the learned parameters of the LLM. Learning the LLM is then usually done by minimizing the negative log-likelihood on a training corpus. Denoting this training data as a multiset\footnote{Oftentimes it is a proper set due to data deduplication~\citep[inter alia]{tirumala2024d4}.} $\data = \{\vy^{(j)}\}_{j=1}^M$, the training criterion is:
\begin{align}
    \mathcal{L}(\text{\vparam}; \data) 
    &= -\sum_{\vy \in \data} \sum_{i = 1}^{|\vy|} w_i(\vy) \log p_{\text{\vparam}}(y_i\mid\vy_{<i}),
    \label{eq:log_likelihood}
\end{align} 
where $w_i(\vy) \geq 0$ are weights that are given to each token\footnote{We will drop the dependency on $\vy$ in the following for notational convenience.}.
This training criterion is also referred to as cross-entropy loss because it minimizes the (weighted) cross-entropy between the model and the empirical data distribution.

\paragraph{Perplexity} \label{para:perplexity} 
The cross-entropy criterion is directly related to the perplexity of a model on a given corpus $\data$, which is defined as the exponential of the average loss over the data with weights $w_i = 1$.
Accordingly, perplexity is often used for assessing the goodness of fit of the model. In contrast, surprisal theory from psycholinguistics shows that humans process words highly non-uniformly \cite{hale_probabilistic_2001}. 

For LLMs, the average perplexity per token position already decreases with increasing token position \cite{gao_pile_2020, reid_gemini_2024} and perplexity can even be negatively correlated with long-range reasoning abilities \cite{levy_same_2024}. This naturally leads to the question if perplexity is the correct measure to evaluate long-context understanding, which is also discussed in~\citet{hu_can_2024}. On the other hand, \cite{chen_long_2024} suggest that the root of the problem may be not so much with perplexity itself but rather that the data does not exhibit enough long-range dependencies (LRDs).
In this work, we focus on evaluating on data that has been specifically collected for long-range dependency modeling to overcome this.

\subsection{Dataset Curation}
\label{subsec:dataset curation}
In most cases, the weights $w_i$ in \cref{eq:log_likelihood} are chosen to be uniformly $1$ such that each token in the sequence receives the same importance. 
In the limit of infinite data, uniform weights are preferred because then the true data-generating distribution is the optimum of the training criterion. 
However, in practice, training data is finite and accordingly a lot of the gains of modern LLMs can be attributed to better choices of the training set $\mathcal{D}$. For example, it has become standard to also train on code to improve the model's reasoning abilities \cite{ma2024at} or use clean LLM-generated text as training data~\citep[inter alia]{gunasekar_textbooks_2023}. 
For long-context understanding specifically, synthetic long-range dependencies have been added to the training data~\citep{wu_long_2024}. 
This is directly connected to using non-uniform weights $w_i$: tokens with weight $w_i = 0$ are ignored, tokens with $w_i \in (0,1)$ downsampled, and tokens with $w_i > 1$ upsampled.
\paragraph{Reinforcement Learning}
By applying Jensen's inequality to Eq. (\ref{eq:log_likelihood}) we obtain \begin{align*}
    \mathcal{L}(\theta; \data) &\geq -\log \sum_{\vy \in \data} \sum_{i = 1}^{|\vy|} w_i(\vy) p_{\text{\vparam}}(y_i\mid\vy_{<i}) \\ &\approx -\log \mathbb{E}_\theta(W)
\end{align*}
with weighting function $W(y_i, \vy_{<i}) = w_i(\vy)$. Now $W$ can be interpreted as the reward function of a reinforcement learning problem with action space $\mathcal{V}$ and state space $\bigcup_{i = 0}^{N}\mathcal{V}^i$ where our goal is to maximize the expected return. In this way, we can see our generalized cross-entropy loss from Eq. (\ref{eq:log_likelihood}) as a form of reward-weighted regression (RWR; \citealp{peters_reinforcement_2007}). RWR was recently applied as a model-based dataset curation technique to pretrain LLMs with document-level \cite{wettig_qurating_2024} and segment-level \cite{korbak_pretraining_2023} weights. We discuss concrete weighting functions in the following and propose new methods in~\cref{sec:methodology}. 
\subsection{Token Weighting}
\label{subsec: loss weighting}
Different weighted cross-entropy criteria have been proposed in the literature. For example, focal loss weighs each class by its confidence, i.e. the predicted class probability~\citep{lin_focal_2018}. 
Related to this, MiLe \citep{su_mile_2024} weighs tokens in language modeling according to one minus the entropy of the token-level model distribution. 
As entropy is a measurement of uncertainty, this weighs uncertain tokens higher, but in doing so runs the risk of excessively favoring tokens that are simply hard to learn or ambiguous, for example, due to plausible variation~\citep[inter alia]{baan-etal-2024-interpreting}.

Empirically, it is also observed that such tokens remain challenging throughout training and~\citet{lin_rho-1_2024} propose $\rho$ which promises to tackle this by using sparse $w_i$.
In particular, $\rho$ builds on~\citet{mindermann_prioritized_2022}, who propose a function that can be used to select data that is learnable but not yet learned. 
For this, $\rho$ relies on training a model checkpoint $\vparam_0$ on a clean dataset $\data_0$. For training the target model $\vparam$ on noisy $\data$, the checkpoint  $\vparam_0$ is then used to filter out tokens whose weights $w_i$ are set to $0$.
Formally, the weights are determined by \begin{align}
\tilde{w}_i^{\rho} &= -\log p_{\text{\vparam}}(i) - \left(-\log p_{\text{\vparam}_0}(i)\right) \\ &= \log\left(\frac{p_{\text{\vparam}_0}(i)}{p_{\text{\vparam}}(i)}\right), \label{eq:MS-rho-dense}
\end{align}
where we will use the shorthand $p_{\text{\vparam}}(i) \coloneq p_{\text{\vparam}}(\vy_i|\vy_{<i})$ for notational simplicity from now on. 
We call this stage {\bf token scoring}.
It can be shown theoretically that this form of token scoring approximately selects optimal examples for reducing test loss~\citep{mindermann_prioritized_2022}. Note that this requires training a second model which can be prohibitively expensive.
A final stage is then {\bf postprocessing} which is applied after token scoring.
In $\rho$, weights are set to $0$ or $\nicefrac{1}{\kappa}$ according to whether they score low or high such that a quantile $1-\kappa \in (0,1]$ is set to $0$, where $\kappa$ is the sparsity ratio. This quantile-based postprocessing is naturally robust to outliers produced by the scoring function. 
Yet, sparsification can also have drawbacks, since part of the signal is ignored. Furthermore, it is not clear whether the choice of $\kappa$ will be robust across datasets and even sequences in the same dataset, which might exhibit different signal-to-noise ratios. 
In the following, we will explore such token-weighting methods thoroughly with the focus of long-context understanding and propose new ones that could overcome the mentioned issues.

\section{Methodology}
\label{sec:methodology}
\begin{table}[]
    \centering
    \resizebox{.95\linewidth}{!}{\begin{tabular}{cc|lc|r}
    \hline
        \multicolumn{2}{c|}{Model Confidence} &  & & Example \\
        Long-ctxt. & Short-ctxt. & Token & Score & Tab.\ref{tab:Case-Study}\\ \hline
        $\Uparrow$ & $\Uparrow$ & Easy & $\Downarrow$ & "ades" \\
        $\Uparrow$ & $\Downarrow$ & LRD & $\Uparrow$ & "Mel" \\
        $\Downarrow$ & $\Uparrow$ & Learnable & $\Uparrow$ & "inspired" \\
        $\Downarrow$ & $\Downarrow$ & Hard & $\Downarrow$ & "one"  \\ \hline
    \end{tabular}}
    \caption{We obtain token weights by contrasting short and long-context models. Here, tokens exhibit different properties: they are either easy, hard, learnable\protect\footnotemark or a long-range dependency (LRD).}
    \label{tab:desiderata}
\end{table}
\footnotetext{The postulate of learnable tokens might seem problematic in cases where the extended context should make the model more uncertain (e.g. by contradicting the previous context), but it has to hold on average. See \cref{sec:app:proofs} for a proof.}
\begin{table*}[]
    \centering
    \resizebox{\linewidth}{!}{\begin{tabular}
    {ll|c|ccc|ccccc|cc}
     Dataset$\rightarrow$& & \multicolumn{1}{c|}{MMLU} & \multicolumn{3}{c|}{Longbench} & \multicolumn{5}{c|}{RULER} &\multicolumn{2}{c}{Average} \\
     Weighting$\downarrow$ & Subset$\rightarrow$ & full & full & $\leq$8k & >8k & combined & 8k & 16k & 24k & 32k & long & full \\
    \multirow{3}{*}{No-train} & 8k & \textit{65.39} &
    35.62 &  -  & -  &
    - &  91.18 & -  &  -  &  - & 
    - & -  \\
     \cline{2-13}
      & 32k & 62.25 &
    38.22 &  39.04  & 37.20  &
    77.87 &  88.56 & 81.91  &  74.40  &  66.60 &
      58.04 & 59.45   \\
    \hline 
     - & LCDE & 63.68  &
     37.43 & 39.57 & 35.34 &
     88.80 & 91.79	&90.83	&87.81&	84.77&	
     63.12& 63.30 \\
     \cline{2-13}
     - & Uniform & 63.44 &
     44.37 & 45.91 & 41.07 &
     89.87&91.82	&91.41&	88.87&	87.38	&
     67.12& 65.89  \\
     \hline
     \multirow{3}{*}{Unfrozen} & Sparse & 60.24  &
     \textbf{46.48} & 46.67 & \textbf{43.69} &
     \textbf{90.42}&\textbf{91.93}	&\textbf{91.68}	&89.40	&88.69	&
      \textbf{68.45}& 65.71 \\
     & Dense & 62.92 & 
     45.09 & \textbf{47.02} & 43.19 &
     90.07&91.40& 91.38	& 89.74	&87.74	&
     67.57 &66.02 \\ \hline
     \multirow{3}{*}{Frozen} & Sparse & \textbf{63.73} & 
     45.09 & 46.35 & 41.87&
     90.37&91.65	&91.50&	89.50	&\textbf{88.84}&	
     67.73 & \textbf{66.40} \\ 
     & Dense & 63.08  &
     43.89 & 46.54 & 41.58 &
     90.34&91.58	&91.63	&\textbf{89.90}&	88.23&	
     67.11&65.77 \\ \hline
    \end{tabular}}
    \caption{ 
    Main results for Llama-3 8B on MMLU, Longbench and RULER (higher is better). The largest value of each column is bolded (values in italic are not considered). While Unfrozen Sparse dominates the 8-16k context, Frozen Sparse has the best performance overall. BBH results are in Table A.\ref{app:tab:bbh} but highly correlated with MMLU.}
    \label{tab:main_results}
\end{table*}

Our goal is to provide an in-depth comparison of existing and novel token-weighting methods for training language models for long-context understanding. 
For this, we present a two-step framework to determine these weights, which allows deriving existing approaches as well as finding new ones. 
The framework consists of the steps \textbf{token scoring} and \textbf{postprocessing}. 
This naturally generalizes the weight calculation and normalization steps from \cite{ren_learning_2019}. 
In the following, we discuss different choices for both steps but first look at the general setting.

We assume to have a language model $\text{\vparam}_0$ that was pre-trained on a dataset $\data_{:n}$ consisting of sequences of length at most $n$ and therefore has context size $n$.
Next, we extend the context size of $\text{\vparam}_0$ to handle sequences of length $N >> n$. 
For example, we can interpolate positions \cite{chen_extending_2023} or increase the base of RoPE embeddings \cite{xiong_effective_2023}. 
The resulting new model $\vparam$ is then subsequently trained on $\data_{:N}$.

Our goal now is to introduce methods that will make the training on $\data_{:N}$ effective for long-range dependencies. 
For this, it is plausible to not restrict weights to sparse $0$s and $1$s but rather to allow any (non-negative) real number as token weight.

\subsection{Token Scoring}
\label{para:token-scoring}
A meaningful token scoring method for context extension should exhibit four desiderata corresponding to four different cases. 
We outline these in \cref{tab:desiderata}. 
All token scoring methods use two models: a long-context and short-context model. 
The latter is oftentimes $\vparam_0$ if context extension is used and the long-context one is a finetuning $\vparam$ of it.
The goal is to score such tokens highly, where the long-context model is certain and the short-context model is uncertain and vice versa, because those either exhibit a long-range dependency (e.g. "Mel" in Tab. \ref{tab:Case-Study}) or are tokens the long-context model has not yet learned but can learn (e.g. "inspired"). Tokens that are inherently hard to predict (e.g. "one") or trivial (e.g. "ades"), however, should get a low weight.
Importantly, these properties change during training, since the long-context model will improve and the weighting of a single token is highly context dependent.
In principle, any model $\vparam^\prime$ can be used for the short-context model, not just $\vparam_0$, for example, a smaller language model for more efficient scoring.

In accordance with these requirements, we choose the following scoring function: \begin{align}
    |\tilde{w}_i| &= 
    \left\lvert
    \log\left( \frac{p_{\text{\vparam}'}^{(n)}(i)}{p_{\text{\vparam}}^{(N)}(i)} \right)
    \right\rvert \\
    &= 
     \left\lvert
    \log( {p_{\text{\vparam}'}^{(n)}(i)})-\log({p_{\text{\vparam}}^{(N)}(i)} )
    \right\rvert.
    \label{eq:pre-normalized}
\end{align}
Here, the superscript indicates how many past tokens the model can use to predict the current token. 
By comparing to Tab. \ref{tab:Case-Study} again, it is easy to see that this fulfills our desiderata: If the numerator is large and the denominator small, we will get a high weight.
If both are roughly equal, the weight is close to 0. 
If the denominator is large and the numerator small, the weight is again large due to the absolute value.

This absolute value also distinguishes our method from the scoring used in $\rho$ shown in \cref{eq:MS-rho-dense}.
Without it, tokens where the short-context model is less certain than the long-context model would get a negative score. This makes sense when both models have the same context size. 
However, when doing context extension, it would violate the second requirement which pushes the model to learn long-range dependencies by upweighing them.
Note that no backpropagation is done through the token weights. Rather, they are treated as constants. 

The ratio $\tilde{w}_i$ is the negative conditional pointwise mutual information (CPMI) between the current token and the faraway context\footnote{A proof can be found in \cref{sec:app:proofs}.}. This CPMI value is an indicator of long-range dependency because it measures the influence of the faraway context on the current token. Therefore it is a natural choice to realize requirement two. 

Note that there are many scoring methods fulfilling the desiderata but ours is arguably one of the simplest and thus should be preferred according to Occam's razor. Empirically, we ablate the importance of the absolute value in Section \ref{sec:ablations}.
Next, we discuss different choices of the base model ${\text{\vparam}^\prime}$ which we use to determine the score.

\paragraph{Choice of base model}
There are various intuitive choices of the base model. The first is to set $\text{\vparam}^\prime = \text{\vparam}_0$ and freeze the short-context model which is used to initialize the long-context model.  
This is similar to the approach of $\rho$. 
However, it either requires scoring the full dataset before training, which is time-consuming, or keeping two models during training in order to perform token scoring which doubles GPU memory consumption.

Intuitively, this could be overcome by sharing the weights of long-context and short-context model, i.e. setting $\text{\vparam}^\prime = \text{\vparam}$, and artificially shortening the context of the model to be trained. 
This way the GPU memory consumption is the same and improvements in the short-context setting might be beneficial for long-context training.
We provide a detailed comparison of both approaches in~\cref{sec:frozen_unfrozen}.

Another approach is to use a smaller model than the long-context model as a short-context model.
The small model is then used as a ``teacher'' similar to weak-to-strong generalization~\citep{burns24weaktostrong}.
For example, current open-source LLMs often come in various sizes with the same base architecture and can be easily used as long as they share the same vocabulary~\citep{dubey_llama_2024, gemma_team_gemma_2024}.

\subsection{Postprocessing}
One option that is used in $\rho$ is to set the weights for the smallest $(1-\kappa)\%$ scores to zero. 
The others are all set to $1/\kappa$ such that they sum to $N$.
This is robust to outliers but might ignore the signal of meaningful tokens with zero weight. Further, due to inherent constraints in autoregressive language modeling with transformer decoders, no speed-up of the backward pass is possible, even if only a fraction of tokens contribute to the loss.
Thus, we employ a dense weighting scheme by using the scores $|\tilde{w}_i|$ directly which provides a dense weighting.
As the scores are unbounded, it is important to \textbf{normalize} them such that they sum to $N$, i.e. \begin{align}
    \text{norm}(|\tilde{w}_i|) = N \cdot \frac{|\tilde{w}_i|}{\sum_{i = 1}^N |\tilde{w}_i|}.
\end{align}
This ensures that the same learning rate can be used as with standard training~\cite{ren_learning_2019}. 
Based on initial experiments we also \textbf{interpolate} them with the standard uniform weighting scheme \begin{align}
    w_i = \lambda + (1-\lambda) \cdot \text{norm}(|\tilde{w}_i|) \label{eq:interpolation}.
\end{align}
Here, $\lambda \in [0,1]$ is a hyperparameter that allows to control how close the weights are to uniform, similar to introducing a temperature parameter.
For $\lambda = 1$ we recover the standard uniform weighting with zero standard deviation, whereas $\lambda = 0$ uses the normalized scores directly, which maximizes temperature. 
Note that interpolation does not affect the normalization and the $w_i$ still sum up to $N$.

\section{Experiments}
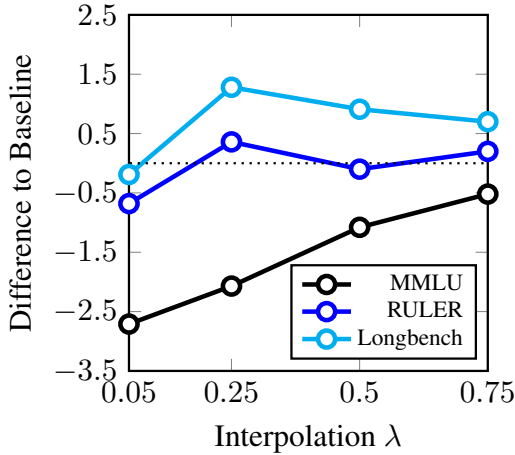
\begin{figure}[t]
\centering
\resizebox{.9\columnwidth}{!}{\begin{tikzpicture}
\begin{axis}[
    xlabel=Interpolation $\lambda$,
    ylabel=Difference to Baseline,
    xmin=0.05, xmax=0.75,
    ymin=-3.5, ymax=2.5,
    height=6cm,
    width=6cm,
    xtick={0.05, 0.25, 0.50, 0.75},
    tick label style={/pgf/number format/fixed},
    ytick={-3.5,-2.5,...,2.5},
    style=thick,
    xlabel near ticks,
    ylabel near ticks,
    legend style={at={(0.97,0.30)}, nodes={scale=0.75, transform shape}},
    legend cell align={right}
            ]
\addplot+[ultra thick,mark=*,black, mark size=3pt, mark options={scale=1, fill=white}] plot coordinates {
(0.05, -2.71)
(0.25, -2.07)
(0.5, -1.08)
(0.75,-0.52)
};
\addplot+[ultra thick,mark=*,blue, mark size=3pt, mark options={scale=1, fill=white}] plot coordinates {
(0.05, -0.68)
(0.25, 0.36)
(0.5, -0.10)
(0.75, 0.2)
};
\addplot+[ultra thick,mark=*,Cyan, mark size=3pt, mark options={scale=1, fill=white}] plot coordinates {
(0.05, -0.19)
(0.25, 1.28)
(0.5, 0.91)
(0.75, 0.70)
};
\addplot[mark=none, black, dotted] {0.0};

\addlegendentry{MMLU}
\addlegendentry{RULER}
\addlegendentry{Longbench}

\end{axis}
\end{tikzpicture}}
    \caption{We compare different $\lambda$ for the \textit{dense} unfrozen setting with Llama-3 8B (\cref{eq:interpolation}). $\lambda=1$ coincides with the baseline (dotted). While short-context benefits from large $\lambda$, for long-context $\lambda = 0.25$ is best.
    }
    \label{fig:abl:interp}
\end{figure}
\begin{figure}[t]
\centering
\resizebox{.9\columnwidth}{!}{\begin{tikzpicture}
\begin{axis}[
    xlabel=Sparsification $\kappa$,
    ylabel=Difference to Baseline,
    xmin=0.20, xmax=0.8,
    ymin=-3.5, ymax=2.5,
    height=6cm,
    width=6cm,
    xtick={0.2, 0.4, 0.6, 0.8},
    tick label style={/pgf/number format/fixed},
    ytick={-3.5,-2.5,...,2.5},
    style=thick,
    xlabel near ticks,
    ylabel near ticks,
    legend style={at={(0.95,0.95)}, nodes={scale=0.75, transform shape}},
    legend cell align={right}
            ]
\addplot+[ultra thick,mark=*,black, mark size=3pt, mark options={scale=1, fill=white}] plot coordinates {
(0.2, -3.2)
(0.4, -2.54)
(0.6, -1.67)
(0.8, -0.79) 
};
\addplot+[ultra thick,mark=*,blue, mark size=3pt, mark options={scale=1, fill=white}] plot coordinates {
(0.2, 0.55)
(0.4, 0.39)
(0.6, -0.15)
(0.8, -1.01)
};
\addplot+[ultra thick,mark=*,Cyan, mark size=3pt, mark options={scale=1, fill=white}] plot coordinates {
(0.2, 2.11)
(0.4, 0.42)
(0.6, -0.22)
(0.8, -0.61)
};
\addplot[mark=none, black, dotted] {0.0};

\addlegendentry{MMLU}
\addlegendentry{RULER}
\addlegendentry{Longbench}

\end{axis}
\end{tikzpicture}}
    \caption{We compare sparsity values $\kappa$ for the \textit{sparse} unfrozen setting with Llama-3 8B (\cref{subsec: loss weighting}). $\kappa=1$ coincides with the baseline (dotted). Increasing $\kappa$ benefits short-context tasks but hurts long-context capabilities. 
    }
    \label{fig:abl:quantile}
\end{figure}
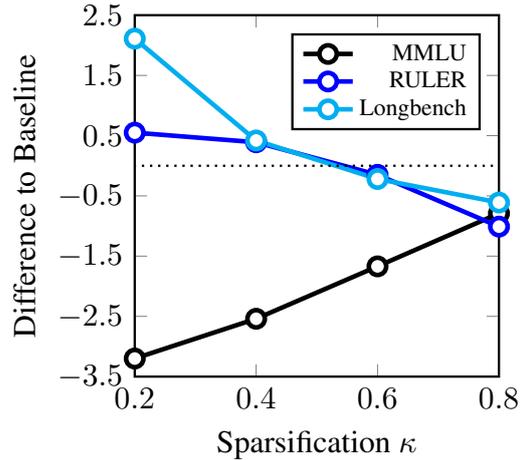
Here we describe the experimental details. 
We first describe the used models, then the datasets which we use for evaluation and finally the used metrics.

\paragraph{Models}
We conduct our experiments on the open-source Llama-3 8B model \cite{dubey_llama_2024}.
Llama-3 8B was trained with a context of 8192 tokens using RoPE embeddings \cite{su2024roformer}.
To extend the context size to 32768 tokens, we increase the base of RoPE from 500k to 15.3M, following Gradient AI\footnote{\url{https://huggingface.co/gradientai/Llama-3-8B-Instruct-Gradient-1048k}}. In initial experiments, this yielded better long-context performance than using 3.5M as the base of RoPE, which is the value predicted by NTK-theory \cite{peng_yarn_2023}.
More detailed information regarding the exact experimental setup, including all technical hyperparameters and runtime comparisons, can be found in \cref{app:sec:exp-details}. 

\paragraph{Weighting variants}
We use four different weighting variants. The Uniform baseline uses standard cross-entropy loss with weights $1$. The \textit{frozen} model uses Llama-3 8B out-of-the-box with 8k context as token scorer, i.e. $\vparam^\prime = \vparam_0$. \textit{Unfrozen} models are their own scorers, i.e. $\vparam^\prime = \vparam$. A \textit{dense} model uses the normalized scores directly, interpolated with uniform loss. 
\textit{Sparse} models use a threshold to set some token weights to zero. 
We train all four combinations \textbf{Sparse Frozen}, \textbf{Sparse Unfrozen}, \textbf{Dense Frozen}, \textbf{Dense Unfrozen} in the exact same setup and report results in \cref{tab:main_results}. 
Unless otherwise noted, dense models use $\lambda = 0.75$ and sparse models use $\kappa = 0.2$ resp. $0.4$ for the unfrozen resp. frozen variant. Ablations for these choices can be found in \cref{fig:abl:interp} and \cref{fig:abl:quantile}. 

\paragraph{Generalizability} To show generalizability, we use the same hyperparameters to extend the context of Phi-2~\citep{javaheripi2023phi} from 2k to 32k by increasing the base of RoPE from 10k to 500k following \cite{chen_extending_2023} and report results in \cref{app:tab:phi2} in the Appendix. Again, the NTK-predicted value of 1.3M performed subpar in initial experiments. Finally, we conduct a weak-to-strong experiment by using scores obtained from smaller models of the Llama family to extend Llama-3 8B in the \textit{frozen} setting, see \cref{tab:weak2strong} for results. 

\paragraph{Datasets}
We use the book corpus PG-19 \cite{rae_compressive_2019} for continual pretraining and tokenize each document separately by splitting it into chunks of size 32k.
We remove the last chunk if it is smaller than 32k tokens. PG19 was chosen because we can obtain 70k continuous sequences of text from it, each of length 32k (2B tokens in total). As an ablation, we also apply Long-Context Data Engineering (LCDE)~\cite{fu_data_2024} to perform per-source length upsampling of sequences longer than 8k tokens from SlimPajama~\citep{sobolevaslimpajama} and packing them together randomly. An additional ablation with rare 32k sequences (less than 0.1\%) from the recent FineWeb dataset \cite{penedo_fineweb_2024} can be found in Table A.\ref{app:tab:fineweb}.

To calculate the loss with a context of 8k, we unfold the 32k sequence with an overlap of 2k tokens to avoid a loss spike. 
For the frozen variant, we perform two forward passes with doubled batch size using Llama 3 8B without context extension.
For the unfrozen variant, we get the logits for the first 8k tokens from the forward pass with the long context model. The four remaining chunks can be combined and computed with two forward passes.\footnote{We also tried using sliding window attention but did not find this strategy to work well, potentially due to the attention sink problem \cite{xiao_efficient_2023}.}

\paragraph{Evaluation}
As mentioned in \cref{para:perplexity}, evaluating models on their long-context understanding capability is challenging.  
Since both using perplexity and needle-in-a-haystack tasks can be inconclusive~\citep{hsieh_ruler_2024, hu_can_2024}, we instead evaluate on the RULER benchmark \cite{hsieh_ruler_2024}.\footnote{For perplexity evaluation see \cref{tab:perplexity} (Appendix).} RULER is a synthetic dataset and consists of the task categories retrieval, multi-hop tracing, aggregation and question answering.
We calculate the RULER score according to~\citet{hsieh_ruler_2024}\footnote{\url{https://github.com/hsiehjackson/RULER}.}. 
In order to obtain the score, the recall over each of the 13 tasks is calculated.
Then, all task-specific recalls are averaged such that each value reflects 6500 sequences of a given length. The \textit{combined} score is the average over sequence length results 8k, 16k, 24k, 32k for Llama-3 resp. 2k, 4k, 8k, 16k, 32k for Phi-2. A detailed description of RULER can be found in \cref{app:sec:eval-data}.
As a non-synthetic benchmark we choose the English part of Longbench \cite{bai_longbench_2023} which consists of 16 tasks over the six categories.
Following \cite{lu_controlled_2024} we further split each task into sequences of length $\leq8k$ and $>8k$ and report the macro-average over tasks. If documents exceed the sequence length of the evaluated model, we truncate in the middle\footnote{ \url{https://github.com/Leooyii/LCEG/}}.
To measure whether long-context models lose performance on short-context, we do 5-shot evaluation on MMLU \cite{mmlu_hendrycks_21}\footnote{We use the code from \url{https://github.com/EleutherAI/lm-evaluation-harness}}. MMLU is well suited as 99\% of its questions are shorter than 500 tokens and the longest one is around 1000 tokens. 
The whole 5-shot prompt mostly consists of less than 3k tokens. 
To diversify the short-context evaluations, we test the models on BigBench-hard in a 5-shot Chain-of-Thought setting \cite{suzgun_bbh} (see Table \ref{app:tab:bbh} in the Appendix).
We average \textit{RULER-combined} and \textit{Longbench-full} to \textit{Long-Average}. 

\section{Results}
\begin{table}[]
    \centering
    \resizebox{\linewidth}{!}{
    \begin{tabular}{l|rrrr}
      Model    & MMLU & Longbench & RULER & Avg.\\  \hline
PPMI            &\textit{59.27}	&43.79 & 85.26 & 62.77\\ 
NPMI            &\textbf{64.06}	&\textit{40.29} &  \textit{77.71} & \textit{60.69}\\
sPPMI           &62.68	&\textbf{45.79} & 90.29 & 66.25 \\
sNPMI           &63.45	&44.36 & 89.47 & 65.76 \\
abs(PMI)        &63.73	&45.09 &  \textbf{90.37} & \textbf{66.40} \\ \hline
    \end{tabular}
    }
    \caption{Ablation over different PMI variants as scoring functions. Bolded values are highest, italic values lowest per column.}
    \label{tab:pmi}
\end{table}

\begin{figure}
    \centering
    \includegraphics[width=1\linewidth]{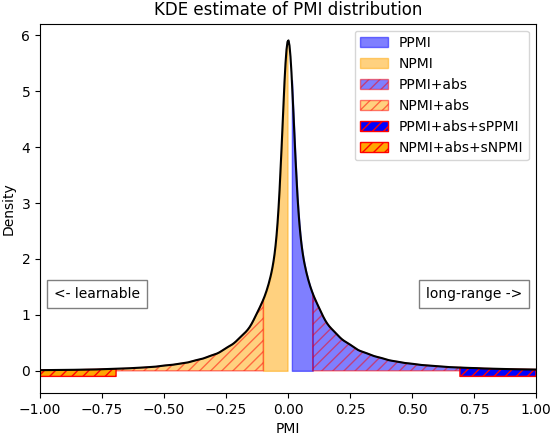}
    \caption{A kernel density estimate of the PMI distribution of all tokens in PG19-eval, measured with the uniform 32k model at the end of training. 
    }
    \label{fig:pmi-kde}
\end{figure}
 In this section, we detail our results from continuously pretraining models with different loss weighting schemes for long-context understanding. 
We first discuss the main results which compare dense and sparse self-weighting models in~\cref{sec:main_results}. 
Then, we discuss the influence of whether the model that is used for weighting tokens is frozen or trained along with the long-context model in~\cref{sec:frozen_unfrozen}.
Finally, we discuss ablations
in~\cref{sec:ablations}.

\subsection{Dense vs. Sparse Weights}
\label{sec:main_results}
In~\cref{tab:main_results} we present results for Llama-3 8B extended to a context of 32k tokens. Sparse unfrozen dominates the 8-16k contexts and is the best model for long context. It should be noted that sequences longer than 16k tokens are not common in Longbench. This performance is remarkable because the sparsity omits 80\% of the tokens from the training. In this way, the model focuses strongly on LRDs which benefits retrieval-heavy tasks. This is further exemplified by Table \ref{appendix:tab:Longbench_by_task}, where Longbench scores are broken down by task. There, the sparse model excels in the QA and synthetic categories. 
But we also see that this model performs worst on MMLU and BBH. The underlying problem is that the unfrozen method cannot score the original 8k tokens in a meaningful way. As there is no context difference, all scores will be zero. Thus, the sparse unfrozen method never considers the first 8k tokens for training. For the dense method, interpolating with standard cross-entropy ensures that all tokens get non-zero weight. Accordingly, the performance of the dense unfrozen model is not as skewed. Although not performing best on any dataset it ranks second on the overall average. To answer \textbf{RQ1: What are the effects of sparse and dense weighting?}, we can say that a sparse weighting pushes the model to perform well on LRDs and specializes it mostly on retrieval-heavy QA tasks. This is because retrieval is a sparse extraction task and the sparse weighting teaches the model to ignore irrelevant information and focus import pieces of information. By choosing a dense weighting on the other hand, the model remains more general and retains much of its short-context ability. 
 
\subsection{Frozen vs. Unfrozen Weighting Model}
\label{sec:frozen_unfrozen}
~\cref{tab:main_results} shows that sparse frozen is the best overall model. It combines the ability of the sparse method to focus on LRDs but can also assign meaningful weights to the first 8k tokens. This is because it uses the frozen Llama 3 8B model as the scorer. Dense frozen is on the level of the uniform weighting for long context and subpar overall. 
Table A.\ref{app:tab:phi2} shows results for Phi-2 2.7B for further insights. First, we see that, immediately after context extension, Phi-2 is very bad on all contexts. Interestingly, this does not harm the unfrozen performance. Apparently, we do not need a good long-context model to start with for the unfrozen approach to work. On the other hand, we see the frozen models falling short here. This might be because here we extend the context by a factor of 16 and not four as for Llama-3. Thus, the model and its frozen scorer might move too far away from each other. 
This hypothesis is supported by the results in Table \ref{tab:weak2strong}, where we use Llama 3.2 1B with 8k context as the frozen scoring model. Here, the results are reasonable with sparse frozen dropping to baseline level and dense frozen even gaining performance.  To answer \textbf{RQ2:
What is the effect of the token scoring?} we can say that, somewhat surprisingly, the unfrozen approach can recover from the model initially being bad on long-context. Additionally, the frozen approach has a hard time coping with differences in context lengths between scorer and model. On the other hand, the size of the scorer does not matter much.

\subsection{Ablations}
\label{sec:ablations}
\begin{table}[]
    \centering
    \resizebox{\linewidth}{!}{
    \begin{tabular}{ll|lll}
      Weighting  & Scoring    & MMLU & Longbench & RULER %
      \\ 
      \hline
      Baseline & -  & 63.44	&	44.37&	89.87 
      \\
      \hline
      Sparse & 3.2 1B& 63.08 {\scriptsize\textcolor{red}{(-0.65)}} & 44.36 {\scriptsize\textcolor{red}{(-0.73)}} & \underline{90.47} {\scriptsize\colorbox{green}{(+0.10)}} \\
      &3.2 3B& 63.39 {\scriptsize\textcolor{red}{(-0.34)}} & \underline{44.78}     {\scriptsize\textcolor{red}{(-0.31)}} & \underline{90.47} {\scriptsize\colorbox{green}{(+0.09)}} \\
      &3.1 8B& 63.31 {\scriptsize\textcolor{red}{(-0.42)}}  & 44.09 {\scriptsize\textcolor{red}{(-1.00)}} & \underline{89.88} {\scriptsize\textcolor{red}{(-0.49)}}
      \\
      \hline
      Dense &3.2 1B & \underline{63.47} {\scriptsize\colorbox{green}{(+0.39)}}  & \underline{44.85} {\scriptsize\colorbox{green}{(+0.96)}} & \underline{90.52}{\scriptsize\colorbox{green}{(+0.18)}}  \\
      
       &3.2 3B & \underline{63.52} {\scriptsize\colorbox{green}{(+0.44)}}  & 44.30 {\scriptsize\colorbox{green}{(+0.41)}} & \underline{90.48}{\scriptsize\colorbox{green}{(+0.14)}}  \\
       &3.1 8B & 63.10 {\scriptsize\colorbox{green}{(+0.01)}}  & \underline{44.71} {\scriptsize\colorbox{green}{(+0.82)}} & \underline{90.24} {\scriptsize\textcolor{red}{(-0.10)}}  \\
      
      \hline
    \end{tabular}
    }
    \caption{Results for weak-to-strong generalization. \textit{Scoring} means the frozen scoring model of the Llama-3.x family. The colours indicate the difference to using the model to be trained (i.e. Llama 3 8B) as frozen scorer. Underlined entries improve over the baseline. Dense models benefit from different scoring models, sparse ones not.
    }
    \label{tab:weak2strong}
\end{table}

Here we discuss several important design choices. 
First we present a dataset comparison and then ablate over interpolation scale and sparsity ratio.

\paragraph{Dataset comparison}
As can be seen in the top rows of Table \ref{tab:main_results}, all our models outperform the LCDE method of \cite{fu_data_2024}, especially on 16k+ context. This makes sense as although their upsampling method creates 32k sequences, manual inspection yields that they mostly consist of \textit{two} random documents, thus limiting the contained LRDs to 16k on average. We again see the problem of "long context is not long at all" exemplified here \cite{chen_long_2024}. On the other hand, training on 32k sequences from FineWeb \cite{penedo_fineweb_2024} yields comparable results to training on PG19 (see Tab. A.\ref{app:tab:fineweb}).

\paragraph{Influence of interpolation scale}
\cref{fig:abl:interp} shows various interpolation parameters $\lambda$ when using a dense weighting and unfrozen Llama 3 model. For long-context data we see the best performance at $\lambda=0.25$.
For MMLU there is a clear trend: as the interpolation value equals the weight of the first 8k tokens, increasing it benefits this task. These gains lead to the maximum average score at $\lambda=0.75$. Thus, we chose $\lambda=0.75$ for our main experiment (Tab. \ref{tab:main_results}). The ablation for dense frozen is more robust and can be found in Figure A.\ref{fig:abl:interp:frozen}.

\paragraph{Influence of sparsity ratio}
When training with sparse weights the hyperparameter $\kappa$ determines how many weights are kept (i.e. $(1-\kappa)\%$ are set to zero) which we refer to as sparsity ratio.
We show an ablation over different $\kappa$ in \cref{fig:abl:quantile} for Llama-3. We see a similar picture as for $\lambda$ but more extreme: the more tokens we use the better for MMLU, but at the same time the focus on LRDs gets blurred. Interestingly, the best average performance is achieved for $\kappa=0.2$. The ablation for sparse frozen is much more robust and can be found in Figure \ref{fig:abl:quantile_frozen}.
\paragraph{Influence of scoring function}
Recall from Section \ref{sec:methodology} that our scoring function can be seen as $abs(\text{PMI})$ where $abs$ is the absolute value and $\text{PMI}$ the (conditional) pointwise mutual information. 
We argued that the absolute value is necessary to fulfill our desiderata. Here, we ablate this choice by comparing to well-known PMI variants. These are $\text{PPMI} :=  \max(\text{PMI},0)$, $\text{NPMI} := \max(-\text{PMI}, 0)$, $\text{sPPMI} := \max(\text{PMI} - \text{ln}(k), 0)$ and $\text{sNPMI} := \max(-\text{PMI} - \text{ln}(k))$. Shifted PPMI with integer shift $k \geq 2$ was introduced by \cite{levy_neural_2014}.
We train our best-performing model, sparse frozen with $\kappa = 0.4$, with these alternative scoring functions and set $k = 2$. The results are in Table \ref{tab:pmi}. We see that PPMI is bad on short context and does also not perform well on long context. If we look at Figure \ref{fig:pmi-kde}, we see that PPMI only selects tokens with high PMI. Some of them are valuable long-range dependencies (high PMI), but we also select a lot of tokens with PMI slightly bigger than zero. As loss decreases naturally with sequence length \cite{gao_pile_2020, reid_gemini_2024}, these tokens are not informative and dilute the training set, leading to suboptimal long context performance. On the other hand, no learnable tokens are selected at all, violating desiderata 3, which leads to bad short-context performance. NPMI, which violates desiderata 2, shows opposite behavior, excelling in short-context but failing to learn long-context. It selects tokens which are learnable but not long-range in Fig. \ref{fig:pmi-kde}. For the shifted variants, roughly 85-95\% of tokens have a score of zero. Thus we randomly sample until we reach $\kappa = 40\%$. Both variants exhibit the same tendencies as their unshifted counterparts, but the random sampling balances out the extremes. Thus, both of them give good overall performance. Fig. \ref{fig:pmi-kde} also shows the superiority of the absolute value, which selects learnable and long-range tokens and avoids uninformative and hard ones. This leads to good overall performance.

\section{Conclusion}
In this work, we show that the weights assigned to tokens in the training objective of LLMs influence their long-context understanding capabilities. 
While standard cross-entropy assigns equal weights to each token, the methods proposed in this work weigh them according to different notions of importance. 
This importance is determined by comparing the confidences of a long- and short-context model.
We compare setting weights densely against having sparse weights and compare using a frozen model for weighting against using a self-weighting approach, where a model assigns its own weights via a forward pass with less context.

Overall, we find that using dense weights improves performance across long-context tasks. Sparse weights mainly improve performance on retrieval-heavy QA tasks, at the cost of short context understanding.
By changing a single hyperparameter we can smoothly trade off long-context and short-context abilities in both approaches.
The frozen weighting scheme is only meaningful if the context is not extended too much but provides a powerful alternative in this setting, by allowing pre-computation and caching of the scores and even using a much smaller model for scoring. 

Our work contributes to a better understanding of token-weighting methods for language models and shows that non-uniform weights can improve long-context abilities. We hope this can spark further investigations into scoring methods, for example, to incorporate uncertainty in other ways.
\section{Limitations}
\label{sec:limitations}
An important limitation of this work is that we did not combine it with other context extension approaches. As we are the first to investigate dense weighting schemes for loss functions, our work is orthogonal to all changes of model architecture and it would be interesting to see with which of the methods it works well together. 
Another limitation is that all our model-based scoring methods need a good language model as a scorer, so our method is not suitable for the early phases of training from scratch. Additionally, we did not instruction-tune our context extended models to keep possible confounders minimal, especially as long-context instruction-tuning is a new area of research in itself \cite{bai_longalign, wu_long_2024}. Studying the interplay of context extension and instruction-tuning with respect to general performance \cite{gao_how_2024} and safety is an exciting direction of future work.
Finally, it remains unclear how our method will scale to modern context lengths of 128k or more. We implemented the sequence parallelism approach of \cite{gao_how_2024} but saw that a single run with 5B tokens takes more than 60 hours on 32 A100 80GB GPUs. This compute demand exceeds our resources, so we will leave scaling questions to future work.
\section{Ethics Statement}
The models presented here are not meant to be deployed directly, because they are not optimized for safety or deployment. Hence, it can not be guaranteed that no harmful content would be generated by these models. 
Apart from that, another ethical consideration is the use of PG19 as a training dataset. Because of copyright issues it only contains novels published before 1919. The societies and values depicted in these books do not hold up to modern standards, which leads the model to inherit these biases. 
\section*{Acknowledgements}
FH is funded by the German Federal Ministry of Education and Research (BMBF) within the hessian.AI Service Center. The model training was supported by a compute grant at the 42 supercomputer provided by hessian.AI (HMD: S-DIW04/0013/003; BMBF: 01IS22091). 
ND is funded by the German Federal Ministry of Education and Research and the Hessian Ministry of Higher Education, Research, Science and the Arts within their joint support of the National Research Center for Applied Cybersecurity ATHENE. 

\bibliography{custom}
\bibliographystyle{acl_natbib}

\appendix

\section{Proofs}
\label{sec:app:proofs}
The following proposition shows the connection between the pointwise mutual information \cite{fano1961transmission} and our token scoring function $\tilde{w}_i$ from equation (\ref{eq:pre-normalized}). The pointwise mutual information measures co-occurrence, i.e. how much more likely it is for two random variables to occur together than we would expect by chance. 
\paragraph{Proposition}
Let $(y_1, \ldots, y_i)$ be a prefix sequence for $\vy \in \mathcal{D}_N$. Split it into the recent context $r_n = (y_{i-n}, \ldots, y_{i-1})$ and the older context $a_{n} = (y_{1}, \ldots, y_{i-(n+1)})$. Then we have \begin{align*}
    -\text{pmi}((y_i, a_{n})|r_n) &:= -\log\left(\frac{p(y_i, a_n|r_n)}{p(y_i|r_n)p(a_n|r_n)}\right) \\ &= \log\left(\frac{p^{(n)}(i)}{p^{(N)}(i)}\right)
\end{align*}
\begin{proof}
We calculate \begin{align*}
    &-\text{pmi}((y_i, a_n)|r_n) \\ &= -\log\left(\frac{p(y_i, a_n | r_n)}{p(y_i|r_n)p(a_n|r_n)}\right) \\ &= -\log\left(\frac{p(y_i, a_n, r_n)p(r_n)^2}{p(r_n) p(y_i, r_n)p(a_n, r_n)}\right) \\ &= -\log\left(\frac{p(y_i, a_n, r_n)p(r_n)}{p(y_i, r_n)p(a_n, r_n)}\right) \\ &= -\log\left(\frac{p(y_i| a_n, r_n)}{p(y_i| r_n)}\right) \\ &= -\log\left(\frac{p(y_i| \vy_{<i})}{p(y_i| r_n)}\right) \\ &=\log\left(\frac{p^{(n)}(i)}{p^{(N)}(i)}\right). 
\end{align*}
\end{proof}
The following Lemma investigates whether it is beneficial for the long-context model to be more certain than the short-context model. It shows that we need a model to be \textit{on average} more certain on the data to fit its underlying distribution better. In particular, while desideratum 3 cannot be fulfilled all the time, we need it to be true on average. \paragraph{Lemma}
Let $\pi, p$ and $q$ be discrete probability distributions. Then we have \begin{align*}
& p(x) \geq q(x) \text{ for all } x \in \text{supp}(\pi)  \\
    \Rightarrow& \mathbb{E}_\pi(\log(p(X))) \geq \mathbb{E}_\pi(\log(q(X))) \\ \iff& KL(\pi, p) \leq KL(\pi, q)
\end{align*} 
\begin{proof}
It holds that \begin{align*}
    \mathbb{E}_\pi(\log(p(X))) &= -CE(\pi, p) \\ &= -KL(\pi, p) - H(\pi)
\end{align*} and analogously for $q$. Applying this to both sides shows the claim.
\end{proof}

\section{Experimental Details}
\label{app:sec:exp-details}
All models are trained on 8 A100-80GB GPUs using Deepspeed Level 3 \cite{rajbhandari2020zero}. 
Following \cite{touvron_llama_2023, chen_extending_2023}, we use a learning rate of $2 \times 10^{-5}$ with 20 linear warmup steps. We train for 240 steps with a batch size of 8 and gradient accumulation of 16 for an effective batch size of 128 (i.e. 4M tokens as in \cite{lu_controlled_2024}). 
We use the AdamW optimizer~\citep{loshchilov2018decoupled} with $\beta_1 = 0.9$, $\beta_2 = 0.95$, $\epsilon = 1e-08$ and weight decay $0.01$.

\subsection{Runtime Analysis}
\label{app:sec:runtimes}
We always used 8 A100 80GB GPUs during training. Preprocessing the whole PG-19 dataset with forward passes for frozen model variants took around 17 hours for Llama 3 8B, 7.5 hours for Phi-2 2.7B and 9.3h for Llama 3.2 1B. Training the baseline or a frozen model for 240 steps took around 30 hours for Llama 3 8B and 15 hours for Phi-2 2.7B . Training the unfrozen model took around 38 hours resp. 17 hours. For a relative runtime comparison see Table \ref{app:tab:runtimes}. We see that from a runtime perspective, using a frozen or unfrozen scoring model makes almost no difference. Scored sequences can be cached and reused in the frozen setting, though. 
\begin{table}[]
    \centering
    \resizebox{\linewidth}{!}{\begin{tabular}{l|lr}
     Model    & Loss & Time factor   \\ \hline 
     Llama 3 8B     &  Uniform & 1.00   \\
        &  Unfrozen & 1.27   \\
        &  Frozen & 1.24   \\ 
        &  Frozen W2S & 1.13   \\ \hline
     Phi 2 2.7B    &  Uniform & 1.00   \\
        &  Unfrozen & 1.13  \\
        &  Frozen & 1.21
    \end{tabular}}
    \caption{Relative model runtimes in comparison to the uniform loss weighting. W2S stands for weak-to-strong generalization.}
    \label{app:tab:runtimes}
\end{table}

\section{RULER Description}
\label{app:sec:eval-data}
For any sufficiently large context length, RULER \cite{hsieh_ruler_2024} creates 500 examples in 13 categories each, so 6500 examples in total. The categories cover four different areas: retrieval, multi-hop tracing, aggregation, and question answering.

\paragraph{Retrieval}
In its most general form, we can describe the retrieval task as follows: There are $m$ key-value pairs hidden in a haystack $h$. At the end we ask for $q$ keys and expect $r$ answers.
\begin{itemize}
\item NIAH: A key-value pair is hidden in a haystack $h$. At the end we ask for the key $k$ and want the value $v$ as the answer.
\begin{itemize}
\item $m = 1$, $k$-type = word, $v$-type = number, $h = 5$ short sentences repeated, $q = 1$, $r = 1$. This is passkey-retrieval \cite{mohtashami_landmark_2023}.
\item $m = 1$, $k$-type = word, $v$-type = number, $h$ = essays, $q = 1$, $r = 1$. This is vanilla NIAH \cite{kamradt_needle_2023}.
\item $m = 1$, $k$-type = word, $v$-type = uuid, $h$ = essays, $q = 1$, $r = 1$
\end{itemize}
    \item NIAH multikey: $m$ key-value pairs are hidden in a haystack $h$. At the end we ask for a single key $k$ and want its value $v$ as the answer. $m = \max$ means that there is no haystack, only key-value pairs.
    \begin{itemize}
        \item $m = 4$, $k$-type = word, $v$-type = number, $h$ = essays, $q = 1$, $r = 1$
        \item $m = \max$, $k$-type = word, $v$-type = number, $h$ = None, $q = 1$, $r = 1$
        \item $m = \max$, $k$-type = uuid, $v$-type = uuid, $h$ = None, $q = 1$, $r = 1$
    \end{itemize}
    \item NIAH multivalue: A single key with $m$ values is hidden in a haystack. At the end we ask for all $m$ values of that key. 
    \begin{itemize}
        \item $m = 4$, $k$-type = word, $v$-type = number, $h$ = essays, $q = 1$, $a = 4$
    \end{itemize}
    \item NIAH multiquery: $m$ key-value pairs are hidden in a haystack. At the end we ask for all $m$ keys and expect $m$ values.
    \begin{itemize}
        \item $m = 4$, $k$-type = word, $v$-type = number, $h$ = essays, $q = 4$, $a = 4$
    \end{itemize}
\end{itemize}
\paragraph{Multi-hop tracing}
This task is a proxy for coreference resolution.
\begin{itemize}
    \item Variable tracking: There are $k$ variable assignments in the haystack, such that they form a chain, i.e. $x_1 = n \ldots x_2 = x_1 \ldots x_k = x_{k-1}$. At the end we ask for all variables with value $n$. \\ $k = 5$, $n$-type: number, $h = 5$ short sentences repeated.
    \end{itemize}  
\paragraph{Aggregation}
This task is a proxy for summarization. A list of synthetic words is constructed. 
\begin{itemize}
    \item common words extraction: Pick 10 words from the list which occur 30 times each, which have to be returned. Generate the haystack by sampling other words from the list thrice each.
    \item frequent words extraction: Sampling from the list follows the zeta distribution. The top 3 most frequent words have to be returned.
    \end{itemize}  
\paragraph{Question Answering}
This task is based on real-world question answering datasets. One document is the needle, the haystack consists of other documents randomly sampled from the same dataset. At the end we ask the question for the needle document.
\begin{itemize}
    \item SQuAD \cite{rajpurkar-etal-2016-squad} is used for single-hop QA.
    \item HotPotQA \cite{yang-etal-2018-hotpotqa} is used for multi-hop QA. The needed documents are not necessarily adjacent.
\end{itemize}    
\paragraph{Prompts} For detailed prompts see Appendix D of \cite{hsieh_ruler_2024}.
\section{Additional Results}
\label{app:sec:additional-results}
\begin{table}[]
    \centering
    \resizebox{\linewidth}{!}{
    \begin{tabular}{l|ccccc}
      Weighting      & MMLU & Longb. & RULER & LAvg. & Avg. \\ \hline
      Uniform   & \textbf{51.48}  & 34.00 & 53.03 & 43.51 &46.17\\
      \hline
      Sparse   & 48.88  & 33.52 & 54.05 &  43.78&45.48\\
      Dense & 50.77  & 33.19 & \textbf{54.92} & \textbf{44.06}
      &\textbf{46.29}\\ \hline
      Sparse frozen &  49.53  & 34.12 & 52.17 & 43.78&45.27\\
      Dense frozen & 50.19  & \textbf{34.26} & 51.81 & 43.03&45.42 \\ \hline
      2k no-train & \textit{56.51}  & 11.99 & - & - &- \\
      32k no-train & 35.24 & 8.64 & 1.87 & 5.26&15.25 \\
      \hline
    \end{tabular}
    }
    \caption{Results for Phi-2 2.7B show that Dense unfrozen dominates performs best on average.}
    \label{app:tab:phi2}
\end{table}

\begin{figure}[t]
\centering
\resizebox{\columnwidth}{!}{\begin{tikzpicture}
\begin{axis}[
    xlabel=Interpolation $\lambda$,
    ylabel=Difference to Baseline,
    xmin=0.00, xmax=0.75,
    ymin=-3.5, ymax=2.5,
    height=6cm,
    width=6cm,
    xtick={0.00, 0.25, 0.50, 0.75},
    tick label style={/pgf/number format/fixed},
    ytick={-3.5,-2.5,...,2.5},
    style=thick,
    xlabel near ticks,
    ylabel near ticks,
    legend style={at={(0.95,0.30)}, nodes={scale=0.75, transform shape}},
    legend cell align={right}
            ]
\addplot+[ultra thick,  mark=*,black, mark size=3pt, mark options={scale=1, fill=white}] plot coordinates {
(0.00, -1.32)
(0.25, -0.7)
(0.5, -0.38)
(0.75,-0.36)
};
\addplot+[ultra thick,mark=*,blue, mark size=3pt, mark options={scale=1, fill=white}] plot coordinates {
(0.00, -0.82)
(0.25, -1.22)
(0.5, 0.36)
(0.75, 0.47)
};
\addplot+[ultra thick, mark=*,Cyan, mark size=3pt, mark options={scale=1, fill=white}] plot coordinates {
(0.00, 0.33)
(0.25,-0.05)
(0.5, -0.85)
(0.75, -0.48)
};

\addplot[mark=none, black, dotted] {0.0};

\addlegendentry{MMLU}
\addlegendentry{RULER}
\addlegendentry{Longbench}

\end{axis}
\end{tikzpicture}}
    \caption{We compare different interpolation values $\lambda$ for the dense \textit{frozen} setting with Llama-3 8B (\cref{eq:interpolation}). $\lambda=1$ coincides with the baseline (dotted). While short-context benefits from large $\lambda$, for long-context it is not clear how to choose $\lambda$.
    }
    \label{fig:abl:interp:frozen}
\end{figure}
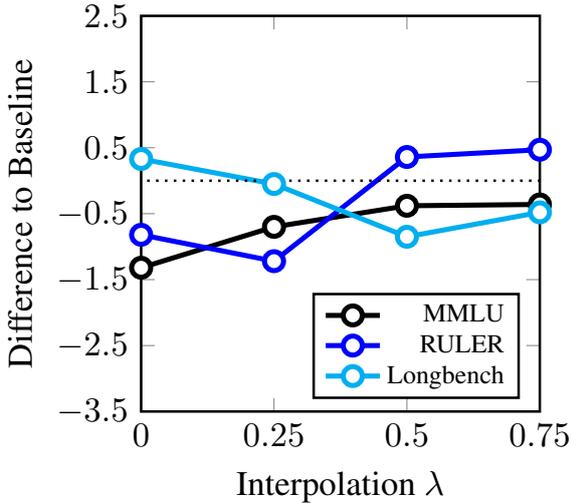
\begin{figure}[t]
\centering
\resizebox{.9\columnwidth}{!}{\begin{tikzpicture}
\begin{axis}[
    xlabel=Sparsification $\kappa$,
    ylabel=Difference to Baseline,
    xmin=0.20, xmax=0.8,
    ymin=-3.5, ymax=2.5,
    height=6cm,
    width=6cm,
    xtick={0.2, 0.4, 0.6, 0.8},
    tick label style={/pgf/number format/fixed},
    ytick={-3.5,-2.5,...,2.5},
    style=thick,
    xlabel near ticks,
    ylabel near ticks,
    legend style={at={(0.95,0.30)}, nodes={scale=0.75, transform shape}},
    legend cell align={right}
            ]
\addplot+[ultra thick,mark=*,black, mark size=3pt, mark options={scale=1, fill=white}] plot coordinates {
(0.2, -0.41)
(0.4, 0.29)
(0.6, -0.16)
(0.8, -0.20) 
};
\addplot+[ultra thick,mark=*,blue, mark size=3pt, mark options={scale=1, fill=white}] plot coordinates {
(0.2, -0.13)
(0.4, 0.5)
(0.6, -0.11)
(0.8, 0.45)
};
\addplot+[ultra thick,mark=*,Cyan, mark size=3pt, mark options={scale=1, fill=white}] plot coordinates {
(0.2, -0.15)
(0.4, 0.72)
(0.6, 0.06)
(0.8, 0.64)
};
\addplot[mark=none, black, dotted] {0.0};
\addlegendentry{MMLU}
\addlegendentry{RULER}
\addlegendentry{Longbench}

\end{axis}
\end{tikzpicture}}
    \caption{We compare different sparsity values $\kappa$ for the sparse \textit{frozen} setting with Llama 3 8B (\cref{subsec: loss weighting}). $\kappa=1$ coincides with the baseline (dotted).  Performance is robust to the choice of $\kappa$. 
    }
    \label{fig:abl:quantile_frozen}
\end{figure}
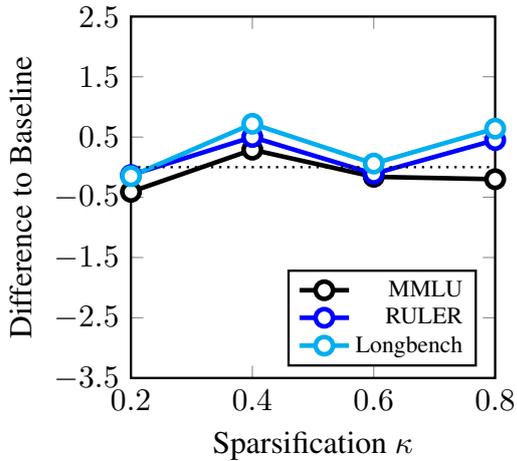
\begin{table*}[]
    \centering
    \resizebox{\linewidth}{!}{\begin{tabular}
    {ll|c|c|c|c|c|c|c}
    \hline
     Weighting$\downarrow$ & Subset$\rightarrow$ & full & Single Doc QA & Multi-Doc QA & Summarization & Few-Shot &  Synthetic & Code   \\
    \hline
    \multirow{3}{*}{No-train} & 8k Baseline & 35.62	&29.30	&27.21	&15.63	&69.04	&4.50	&68.69 \\
     \cline{2-9}
                              & 32k Baseline & 38.22	&31.37	&36.50	&17.56	&68.83	&5.78	&68.58  \\
     \hline
     - & LCDE & 37.43 & 31.32 & 30.75 & 25.01 & 68.92 & 10.25 & 55.20 \\ \hline
                            - & Uniform & 44.37	&35.20	&38.25	&\textbf{27.08}	&\textbf{69.32}	&28.30	&\textbf{71.90}   \\
     \hline
     \multirow{3}{*}{Unfrozen} & Sparse & \textbf{46.48}	&\textbf{35.50}	&\textbf{39.23}	&26.98	&69.15	&\textbf{44.78}	&70.76 \\
     \cline{2-9} 
                              & Dense & 45.07	&34.80&	39.00	&25.94	&68.60	&36.59	&71.46 \\ \hline
     \multirow{3}{*}{Frozen} & Sparse & 45.09	&35.08	&38.62	&27.04	&68.97	&34.53	&71.62 \\ \cline{2-9}
                             & Dense & 43.89	&35.00	&38.10&	25.34&	68.29	&30.07	&70.95
    \end{tabular}}
    \caption{ 
    Results on Longbench by task (higher is better). The largest value of each column is bolded. Most interesting is the Synthetic category, where Sparse Unfrozen excels and Uniform falls short.}
    \label{appendix:tab:Longbench_by_task}
\end{table*}

\begin{table}[]
    \centering
    \resizebox{\linewidth}{!}{
    \begin{tabular}{l|rrrr}
    \hline
      Context    & 8192 & 16384 & 24576 & 32768   \\ \hline
      32k no-train & 10.55 & 10.59 & 10.55 & 10.47 \\
      LCDE & 9.33 & 9.24 & 9.16 & 9.06 \\
      Uniform   &  \textbf{8.83} & \textbf{8.75} & \textbf{8.66} & \textbf{8.55} \\ \hline
      Sparse     & 10.56  & 10.18 & 9.91 &  9.69\\
      Dense      & 9.10  & 9.04 & 8.97 & 8.87\\ \hline
      Sparse frozen   &  9.24  & 9.16 & 9.07 & 8.96\\
      Dense frozen & 8.96  & 8.88 & 8.80 & 8.70

    \end{tabular}
    }
    \caption{Perplexity results for Llama-3 8B on the validation set of PG19. The uniform weighting dominates, but this is not reflected in downstream task performance as shown throughout the paper. }
    \label{tab:perplexity}
\end{table}

\begin{table}[]
    \centering
    \resizebox{\linewidth}{!}{
    \begin{tabular}{l|rr}
    \hline
      Model    & MMLU & BBH \\ \hline
       8k &65.39	&62.20 \\
32k no train&62.25	&53.65 \\ \hline 
LCDE &63.68	&59.24 \\
Uniform&63.44	&58.30 \\ \hline
Unfrozen Sparse &60.24	&52.97 \\
Unfrozen Dense &62.92	&55.83 \\ \hline
Frozen Sparse&63.73	&58.22 \\
Frozen Dense &63.08	&56.04 \\
    \end{tabular}
    }
    \caption{Additional results of Llama 3 8B models on BigBench-hard with 5-shot Chain-of-thought. The performance on both short context datasets is highly correlated with Pearson's $\rho = 0.9289$ and Spearman's $\rho = 0.9286$.}
    \label{app:tab:bbh}
\end{table}

\begin{table}[]
    \centering
    \resizebox{\linewidth}{!}{
    \begin{tabular}{l|lll}
    \hline
      Model    & MMLU & Longbench & RULER \\ \hline
        Uniform& 63.63	{\scriptsize\colorbox{green}{(+0.19)}} &40.02 {\scriptsize\textcolor{red}{(-4.35)}} & 90.54 {\scriptsize\colorbox{green}{(+2.27)}}\\ \hline
Unfrozen Sparse& 59.30	{\scriptsize\textcolor{red}{(-0.94)}} &43.33 {\scriptsize\textcolor{red}{(-3.15)}} & 89.95 {\scriptsize\colorbox{green}{(+2.38)}}\\
Unfrozen Dense & 63.45	{\scriptsize\colorbox{green}{(+0.53)}} &43.32 {\scriptsize\textcolor{red}{(-1.74)}} & 91.26 {\scriptsize\colorbox{green}{(+2.10)}}\\ \hline
Frozen Sparse  & 63.63	{\scriptsize\textcolor{red}{(-0.10)}} &41.77 {\scriptsize\textcolor{red}{(-3.32)}} & 91.38 {\scriptsize\colorbox{green}{(+2.32)}}\\
Frozen Dense   & 63.77	{\scriptsize\colorbox{green}{(+0.72)}} &43.64 {\scriptsize\textcolor{red}{(-0.25)}} & 91.05 {\scriptsize\colorbox{green}{(+2.01)}}\\
    \end{tabular}
    }
    \caption{Additional results of Llama 3 8B models trained on continuous 32k sequences of general pretraining corpus Fine Web \cite{penedo_fineweb_2024}. Sequences of this length amount to less than 0.1\% of the dataset, so heavy filtering is necessary. Noted in brackets is the performance difference towards the standard models trained on PG19 \cite{rae_compressive_2019}. With FineWeb, we get consistent improvements on the synthetic RULER benchmark, but lose even more performance on Longbench, while MMLU performance stays largely constant.}
    \label{app:tab:fineweb}
\end{table}

\end{document}